\documentclass[10pt,twocolumn,letterpaper]{article}

\usepackage{cvpr}
\usepackage{times}
\usepackage{epsfig}
\usepackage{graphicx}
\usepackage{placeins} 
\usepackage{booktabs} 
\usepackage{amsmath}
\usepackage{amssymb}
\usepackage[utf8]{inputenc}

\newcommand{\mat}[1]{\mathbf{#1}}

\newcommand{\ewprod}{\odot}
\newcommand{\reals}{\mathbb{R}}

\newcommand{\bmx}[0]{\begin{bmatrix}}
\newcommand{\emx}[0]{\end{bmatrix}}

\newcommand{\vect}[1]{\mathbf{#1}}
\newcommand{\vects}[1]{\boldsymbol{#1}}

\newcommand{\TT}[0]{\vects{\theta}}

\usepackage[breaklinks=true,bookmarks=false]{hyperref}

\cvprfinalcopy 

\def\cvprPaperID{3167} 

\begin{document}

\title{A dataset and exploration of models for understanding video data through fill-in-the-blank question-answering}

\author{
Tegan Maharaj$^1$ \and Nicolas Ballas$^2$ \and Anna Rohrbach$^3$ \and Aaron Courville$^2$  \and Christopher Pal$^1$ \\
\and
$^1$Polytechnique Montréal \and $^2$Université de Montréal \and $^3$Max-Planck-Institut für Informatik, Saarland Informatics Campus \\ \and
{\tt\small $^1${tegan.maharaj,christopher.pal}@polymtl.ca} \and 
{\tt\small $^2${nicolas.ballas,aaron.courville}@umontreal.ca} \and 
{\tt\small $^3$arohrbach@mpi-inf.mpg.de}
}
\maketitle

\begin{abstract}
While deep convolutional neural networks frequently approach or exceed human-level performance at benchmark tasks involving static images, extending this success to moving images is not straightforward. Having models which can learn to understand video is of interest for many applications, including content recommendation, prediction, summarization, event/object detection and understanding human visual perception, but many domains lack sufficient data to explore and perfect video models.  
In order to address the need for a simple, quantitative benchmark for developing and understanding video, we present MovieFIB, a fill-in-the-blank question-answering dataset with over 300,000 examples, based on descriptive video annotations for the visually impaired. 
In addition to presenting statistics and a description of the dataset, we perform a detailed analysis of 5 different models' predictions, and compare these with human performance. We investigate the relative importance of language, static (2D) visual features, and moving (3D) visual features; the effects of increasing dataset size, the number of frames sampled; and of vocabulary size. We illustrate that: this task is not solvable by a language model alone; our model combining 2D and  3D visual information indeed provides the best result; all models perform significantly worse than human-level. We provide human evaluations for responses given by different models and find that accuracy on the MovieFIB evaluation corresponds well with human judgement. We suggest avenues for improving video models, and hope that the proposed dataset can be useful for measuring and encouraging progress in this very interesting field.
\end{abstract}

\section{Introduction}

\begin{figure}
\center
\includegraphics[width=0.45\textwidth]{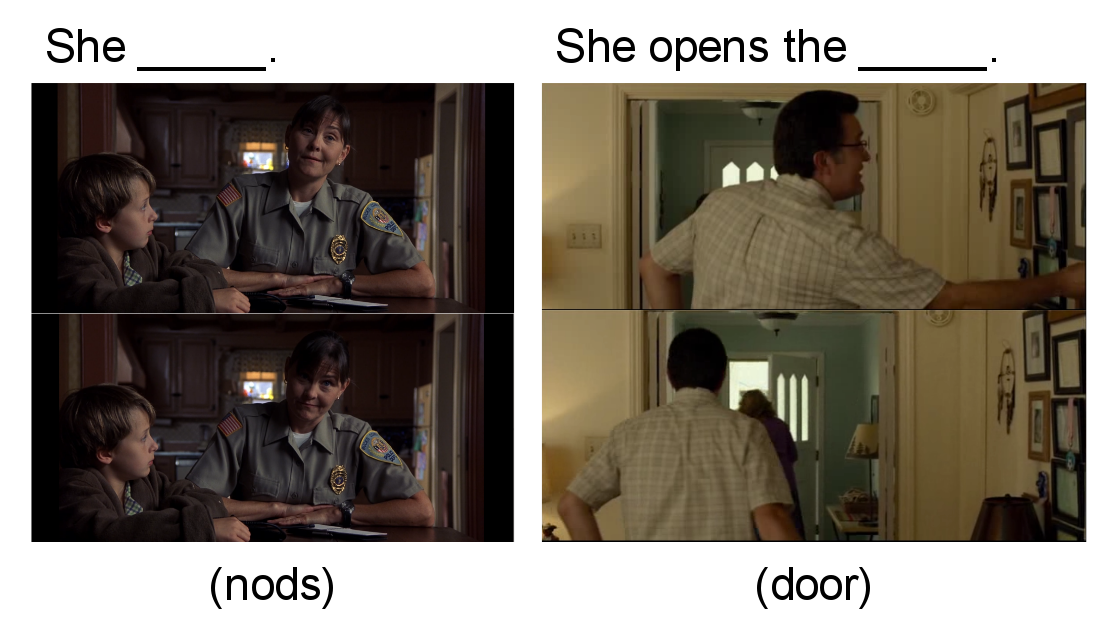}
\caption{Two examples from the training set of our fill-in-the-blank dataset.}
\label{fig:qual_example}
\end{figure}

A long-standing goal in computer vision research is complete understanding of visual scenes: recognizing entities, describing their attributes and their relationships.
The task of automatically translating videos containing rich and open-domain activities into natural language requires tackling the challenges above which still stand as open problems for computer vision.

A key ingredient for sparking the impressive recent progress in object category recognition ~\cite{krizhevsky2012imagenet} has been the development of large scale image recognition datasets for training and evaluation ~\cite{deng2009imagenet}. Accordingly, several large video datasets have been proposed~\cite{rohrbach15cvpr, AtorabiM-VAD2015} to address the video translation to natural language problem.
Those datasets rely on transcriptions of audio narrations from descriptive video services (DVS) (included in movies as an aide for the blind) to obtain text based descriptions of movie scenes.
DVS provides an audio narration of the most important aspects of the visual information relevant to a movie which typically consists of descriptions for human actions, gestures, scenes, and character appearance~\cite{rohrbach15cvpr}.

While the extraction of scene descriptions from DVS has proven to be a reliable way to automatically associate video with text based descriptions, DVS  provides only one textual description per segment of video despite the fact that multiple descriptions for a given scene are often equally applicable and relevant.
This is problematic from an evaluation perspective; standard evaluation metrics used for the video to natural language translation task, such as BLEU, CIDER or Rouge, have been shown to not correlate well with human assessment when few target descriptions are available ~\cite{papineni2002bleu, lin2004rouge, vedantam2015cider}. 
Therefore, it is questionable to rely on such automated metrics to evaluate and compare different approaches on those datasets.

To address these issues, we propose recasting the video description problem as a more straightforward classification task by reformulating descriptions as a fill-in-the-blank question-answering problem.  Specifically, given a video and its description with one word blanked-out, our goal is to predict the missing word as illustrated in Figure~\ref{fig:qual_example}.
Our approach to creating fill-in-the-blank questions allows them to be easily generated automatically from a collection of video
descriptions, it does not require extra manual work and can therefore be scaled to a large number of queries.
Through this approach we have created 300,000 fill-in-bank question and video pairs.
The corresponding fill-in the blank questions concern entities, actions and attributes. Answering such questions therefore implies that a model must obtain some level of understanding of the visual content of a scene, as a model would need to be able to detect   objects and people, aspects of their appearance, activities and interactions, as well as features of the general scene context of a video.


In our work presented here we investigate several baseline models for this tasks. In particular, we show that a language model alone is not able to solve these types of question answering tasks and that the best performance can be attained when static (2D) visual features are combined with moving (3D) visual features. We also show that all of our models are significantly worst than human performance -- leaving room for improvement through further technical advances. Finally, we empirically demonstrate that using classification accuracy for this problem is a robust metric to evaluate and compare models on such a task, as it corresponds well with human judgment.

\section{Related work}

\subsection{Video Captioning}

The problem of bridging the gap between video
and natural language has attracted significant recent attention. Early models tackling video captioning such as \cite{kojima2002, rohrbach2013},
focused on constrained domains with limited appearance of activities and objects in videos and depended heavily on hand-crafted video features, followed by a template-based or shallow statistical machine translation. However, recently models such as~\cite{venugopalan2014translating, donahue2014long, yao2015describing, Nicolas15}
have shifted toward a more general encoder-decoder neural approach to tackle the captioning problem for open
domain videos.
In such architecture, videos  are usually encoded into a vector representation using a convolutional neural network, and then fed to a caption decoder usually implemented with a  recurrent neural networks.

The development of encoder decoder type of models have been possible with the release of large scale datasets~\cite{Guadarrama_youtube2text,AtorabiM-VAD2015,rohrbach15cvpr}.
In particular,~\cite{AtorabiM-VAD2015,rohrbach15cvpr} have exploited descriptive video data
to construct captioning datasets that have a large number of video clips.
Indeed, many movies and TV shows are produced with an additional audio track called \textbf{descriptive video (DV)}. This track is a kind of narration designed for the visually impaired; it supplements the ordinary dialogue and audio tracks of the movie by describing the visual content of a scene in detail. This type of description is very appealing for machine learning methods, because the things described tend to be those which are relevant to the plot, but they also stand alone as 'local' descriptions of events and objects with associated visual content.
In \cite{lsmdc2015,AtorabiM-VAD2015,rohrbach15cvpr}, the authors create a dataset by cutting 200 HD Hollywood movies into 128,085 short (4-5 second) clips, and transcribing the DV track to create clip-annotation pairs. This dataset was used as the basis of the Large Scale Movie Description Challenge (LSMDC) in 2015 and 2016 \cite{lsmdc-challenge}.

While the development of those datasets lead to new models that can produce impressive descriptions in terms of their syntactic and semantic quality, the evaluation of such techniques is challenging \cite{lsmdc2015}. Many different descriptions may be valid for a given image and as we have motivated above, commonly used metrics of quality such as BLEU, METEOR, ROUGE-L and CIDEr ~\cite{papineni2002bleu, denkowski2014meteor,lin2004rouge, vedantam2015cider} have been found to correlate poorly with human judgments of description quality and utility \cite{lsmdc2015}.




\subsection{Image and Video QA}

One of the first large scale visual question answering datasets is the visual question answering (VQA) challenge introduced in ~\cite{VQA}. It consists of 254,721 images from the MSCOCO \cite{lin2014microsoft} dataset plus imagery of cartoon-like drawings from an abstract scene dataset \cite{zitnick2016adopting}. There are 3 questions per image for a total of 764,163 questions with 10 ground truth answers per question. There are three plausible answers per question. The challenge includes questions with possible responses of yes, no, or maybe as well as open-ended and free-form questions and answers provided by humans. Amazon Mechanical Turk was used to create both questions and answers. Other work has looked at algorithmically transforming MSCOCO descriptions into question format creating the COCO-QA dataset \cite{ren2015exploring}. The DAtaset for QUestion Answering on Real-world images (DAQUAR) was introduced in \cite{malinowski2014multi}. It was built on top of the NYU-Depth V2 dataset which consists of 1,449 RGBD images \cite{silberman2012indoor}. They collected 12,468 human question-answer pairs focusing on questions involving identifying 894 categories of objects, colors of objects and the number of objects in a scene.

Following this effort, recent work has also examined video QA formulated as a multiple choice fill in the blank problem \cite{zhu2015uncovering}. They used an encoder-decoder RNN architecture for examining the performance of different approaches to solving this problem. Their data is created by reformulating various video description datasets including TACoS \cite{regneri2013grounding}, MPII-MD \cite{rohrbach2015dataset} and the TRECVID MEDTest 14 \cite{TRECVIDMED14} dataset. Since they used a multiple choice format, the selection of possible answers has an important impact on model performance. They generated questions according to two different levels of difficulty by controlling the level of similarity of possible responses to the true answer. To avoid these issues here we work with an open vocabulary fill in the blank format for our video QA formulation.

Other recent work has developed MovieQA, a dataset and evaluation based on a question answering formulation for story comprehension using both video and text resources associated with movies \cite{MovieQA}. The MovieQA dataset is composed of 408 subtitled movies, along with: summaries of the movie from Wikipedia, scripts obtained from the Internet Movie Script Database (IMSDb) -- which are available for almost half of the movies, and descriptive video service (DVS) annotations -- which are available for 60 movies using the MPII-MD \cite{rohrbach2015dataset} DVS annotations.

\begin{table}
  \center
  \caption{Comparison of our fill-in-the-blank (FIB) dataset with the MovieQA dataset, showing the number of movies, FIB query-response examples (note that number of words includes the blank for FIB)}
  \small
  \begin{tabular}{lrrrr}
  \toprule
     \textbf{MovieQA dataset}& Train & Val & Test & Total \\
\midrule
\#Movies &93& 21& 26& 140 \\
\#Clips& 4,385& 1,098 &1,288& 6,771\\
Mean clip dur. (s)& 201.0 &198.5 &211.4& 202.7$\pm$216.2\\
\#QA& 4,318& 886& 1,258& 6,462\\
Mean \#words in Q& 9.3 &9.3& 9.5& 9.3$\pm$3.5\\
\midrule
     \textbf{MovieFIB dataset}& Train & Val & Test & Total \\
\midrule
\#Movies &200& 153& 12& 17 \\
\#Clips& 101,046 &7,408& 10,053 &128,085 \\
Mean clip dur. (s)& 4.9 & 5.2 &4.2 & 4.8\\
\#QR& 296,960& 21,689& 30,349 &348,998\\
Mean \#words in Q& 9.94& 9.75& 8.67& 9.72\\
\bottomrule
  \end{tabular}
\label{tab:movieQA_comparison}
\end{table}


\section{MovieFIB: a fill-in-the-blank question-answering dataset }
\subsection{Creating the dataset} \label{makedataset}
The LSMDC2016 dataset \cite{lsmdc2015,AtorabiM-VAD2015,rohrbach2015dataset} forms the basis of our proposed fill-in-the-blank dataset (MovieFIB) and evaluation. Our procedure to generate a fill-in-the-blank question from an annotation is simple. For each annotation, we use a pretrained maximum-entropy parser \cite{maxentRatnaparkhi1996,maxentNLTK} from the Natural Language Toolkit (NLTK) \cite{nltk} to tag all words in the annotation with their part-of-speech (POS). We keep nouns, verbs, adjectives, and adverbs as candidate blanks, and filter candidates through a manually curated stoplist (see supplementary materials). Finally, we keep only words which occur 50 times or more in the training set. 

\subsection{Dataset statistics and analysis}
The procedure described in section \ref{makedataset} gives us 348,998 examples: an average of 3 per original LSMDC annotation. We refer to the annotation with a blank (e.g. 'She \_\_\_\_\_ her head') as the \textbf{question} sentence, and the word which fills in the blank as the \textbf{answer}.  We follow the training-validation-test split of the LSMDC2016 dataset; 296,960 training, 21,689 validation, and 30,349 test. Validation and test sets come from movies which are disjoint from the training set. We use only the public test set, so as not to provide ground truth for the blind test set used in the captioning challenge). Some examples from the training set are shown in Figure \ref{fig:qual_example}, and Table~\ref{tab:movieQA_comparison} compares statistics of our dataset with the MovieQA dataset. For a more thorough comparison of video-text datasets, see \cite{lsmdc2015} 

\begin{figure}
\center
\includegraphics[width=0.48\textwidth]{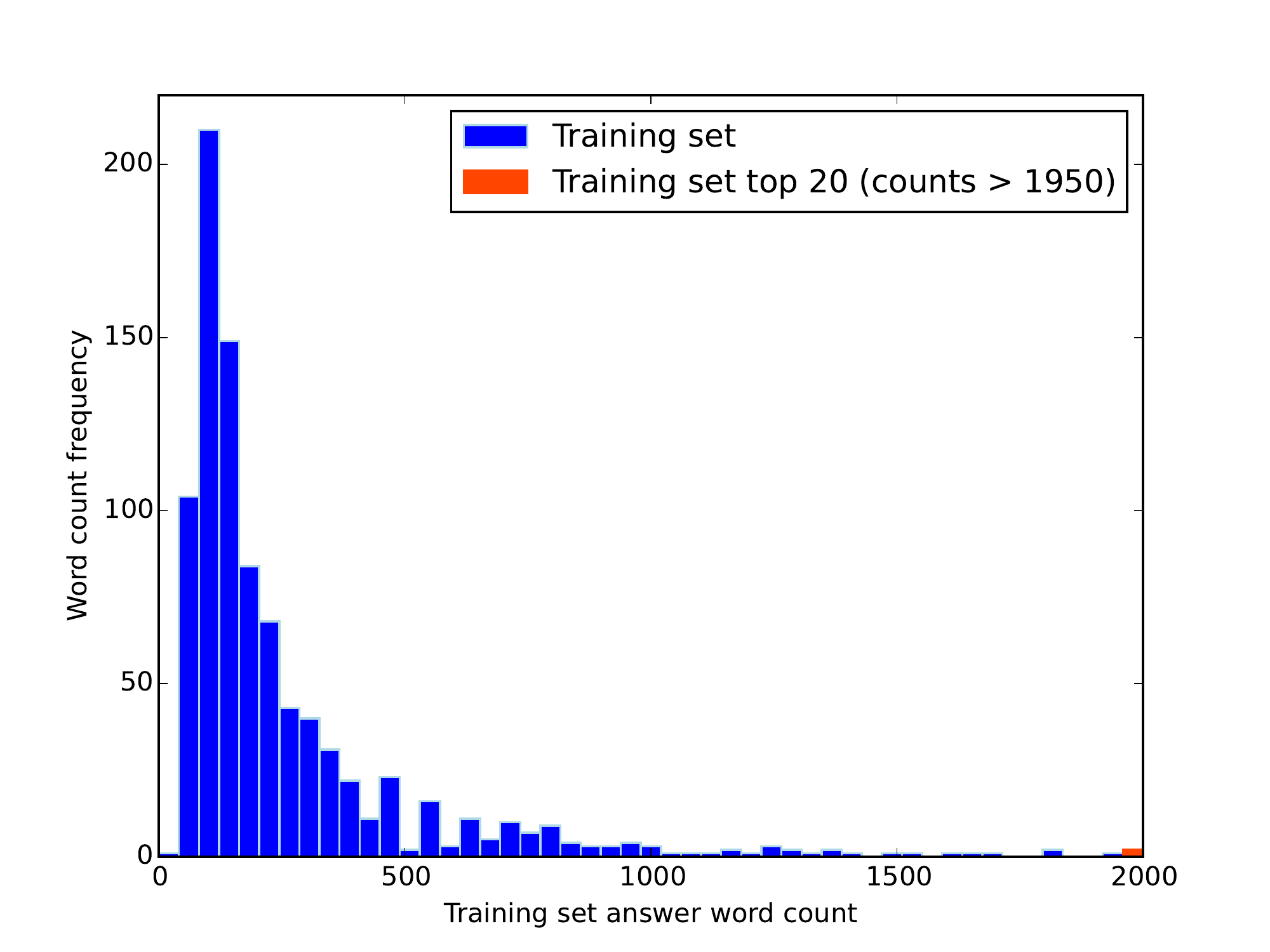}
\caption{Histogram showing frequencies of counts for answers (targets) in the training set. Note that the last bin of the histogram covers the interval [1950 : 12541], containing the 20 most frequent words which are listed in Figure \ref{fig:wordcloud}}
\label{fig:histogram}
\end{figure}

\begin{figure}
\center
\includegraphics[width=0.48\textwidth]{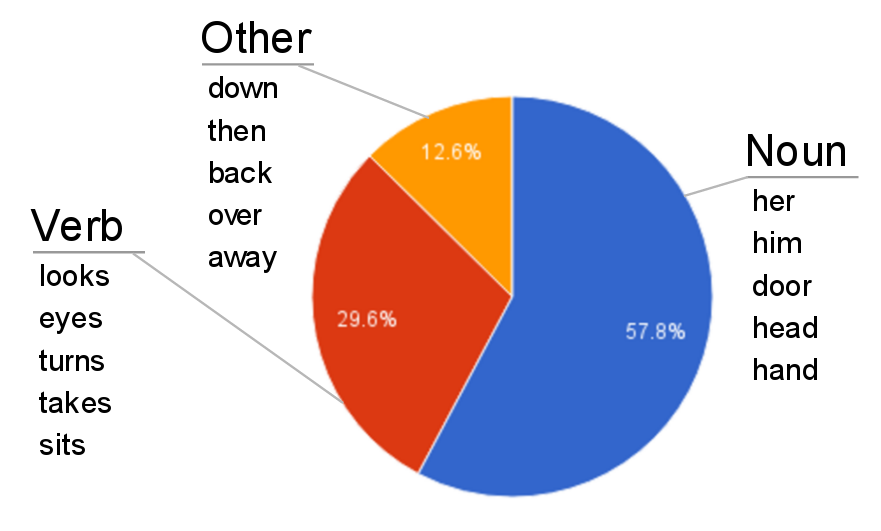}
\caption{Pie chart showing the answer words of the training set by POS-tag category (noun, verb, or other), with the five most frequent words per category}
\label{fig:pospie}
\end{figure}

\begin{figure}
\center
\includegraphics[width=0.48\textwidth]{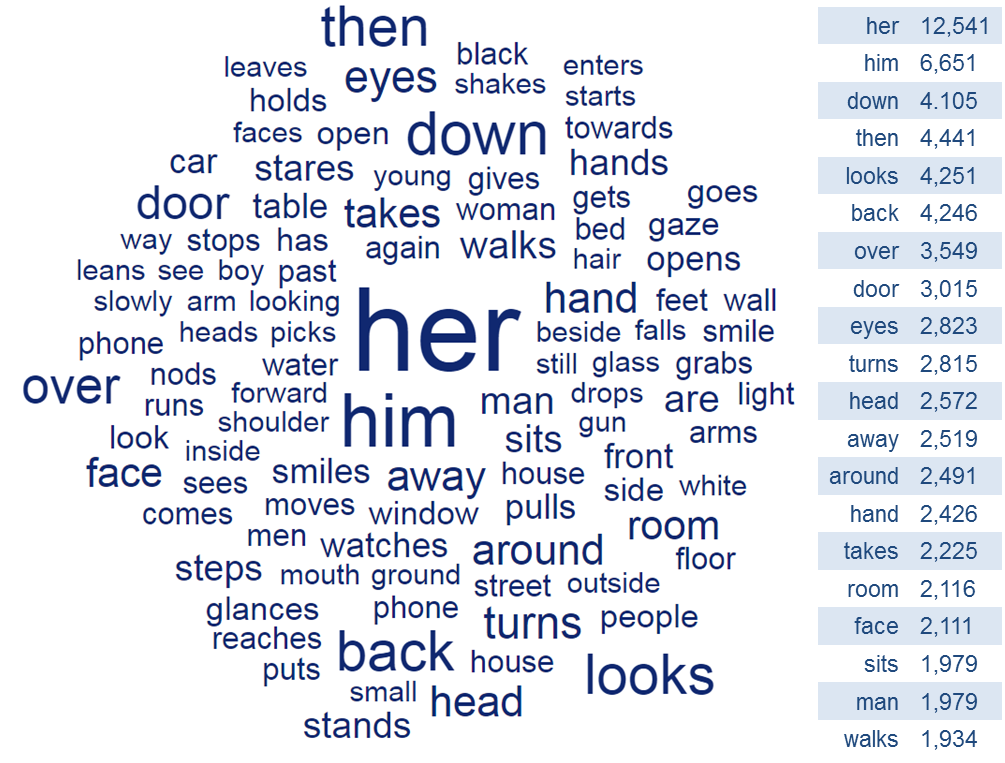}
\caption{Word cloud showing the top 100 most frequently-occurring words in the training set answers (font size scaled by frequency) and list with counts of the 20 most frequent answers}
\label{fig:wordcloud}
\end{figure}

Figure \ref{fig:histogram} is the histogram of  responses (blanked-out words) for the training set, showing that most words occur 100-200 times, with a heavy tail of more frequent words going up to 12,541 for the most frequent word (her). For ease of viewing, we have binned the 20 most frequently-occurring words together. Figure \ref{fig:wordcloud} shows a word-cloud of the top 100 most frequently occurring words, with a list of the most frequent 20 words with their counts. In Figure \ref{fig:pospie} we examine the distribution by POS tag, showing the most frequent words for each of these categories.

\section{Neural framework for video fill-in-the-blank question-answering}
\label{sec:framework}

In this section, we describe a general neural network-based approach  to address fill-in-the-blank video question-answering problems. This neural network provides a basis for all of our baseline models.

We consider a training set $(\vect v^i, \vect q^i, y^i)_{i \in (0..N)}$  with videos $\vect v^i$, questions $\vect q^i$ and their associated answers $y^i$.
Our goal is to learn a model that predicts $y^i$ given $\vect v^i$ and $\vect q^i$.

\begin{figure}
\center
\includegraphics[width=0.40\textwidth]{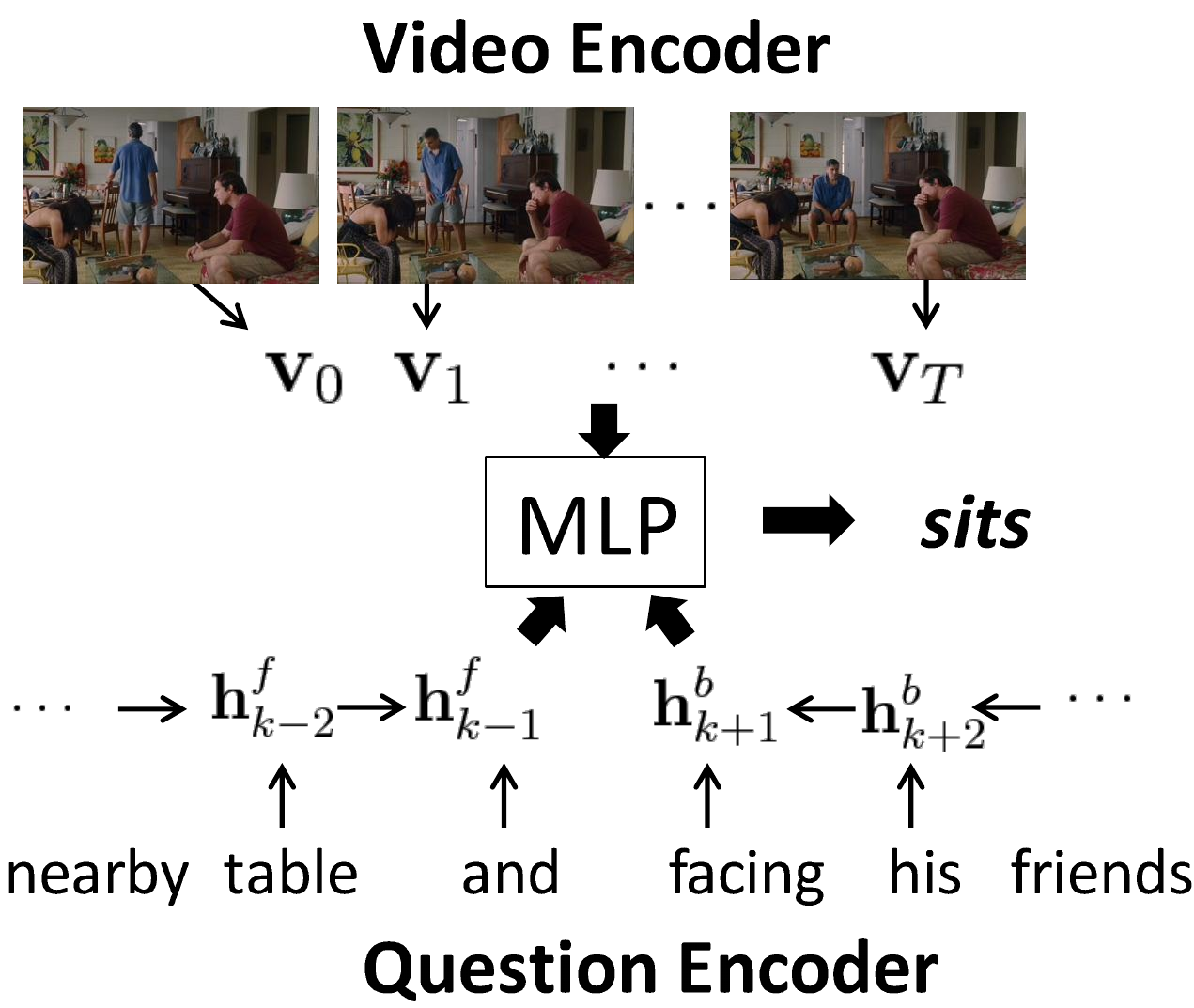}
\caption{Fill-in-the-blank Model Architecture.}
\label{fig:model}
\end{figure}

We first extract fixed length representations from
a video and a question using encoder networks $\Phi_v$ and $\Phi_q$ applied respectively on the video and question  as illustrated in Figure~\ref{fig:model}.
 The fixed length representations are then fed to a classifier network $f$ that outputs a probability distribution over the different answers, $p(y \mid \vect v^i, \vect q^i)) = f(\Phi_v(\vect v^i), \Phi_q(\vect q^i))_y$. $f$ is typically an MLP network that uses a softmax activation function in its last layer.

We estimate the model parameters $\TT$ composed by the encoder and classifiers networks parameters $\TT = \{\TT_v, \TT_q, \TT_f\}$ by maximizing the model log-likelihood
on the training set,
\begin{equation}
\mathcal{L}(\TT) = \frac{1}{N} \sum_{i=1}^N \log p(y^i \mid \vect v^i, \vect q^i), \TT).
\end{equation}

\subsection{Question Encoder}
Recurrent neural networks have become the standard neural approach to encode text, as text data is composed of a variable-length sequence of symbols~\cite{sutskever2014sequence,cho2014learning}. Given a sequence of words $\vect w_t$ composing a question $\vect q$, we define our encoder function as $\vect h_t = \Phi_q(\vect h_{t-1}, \vect w_t)$ with $\vect h_0$ being a learned parameter.

In particular, we are here interested in a fill-in-the-blank task, a question $\vect q$ composed by $l$ words
can therefore be written as $ \vect q = \{ \vect w_0, \ldots \vect w_{k-1}, \vect b, \vect w_{k+1}, \vect w_l \}$, where $\vect b$ is the symbol representing the blanked word.
To exploit this structure, we decompose our encoder $\Phi_q$ in two recurrent networks, one forward RNN $\Phi^f_q$ which will be applied on the sequence $\{ \vect w_0, \ldots, \vect w_{k-1}\}$, and one  backward RNN
that will be applied on the reverse sequence $\{ \vect w_{T}, \ldots, \vect w_{k+1}\}$.
The foward hidden state $\vect h^f_{k-1}$ and backward hidden state $\vect h^b_{k+1}$ are then concatenated
and provided as input to classifier networks. Similar network structure for fill-in-the-blank question has also been explored in~\cite{mazaheri2016video}.

Forward and backward functions $\Phi^f_q$ and  $\Phi^b_q$ could be implemented using vanilla RNNs,
however training such model using stochastic gradient descent is notoriously difficult
due to the exploding/vanishing gradients problems~\cite{bengio1994learning,hochreiter1991untersuchungen}.
Although solving those gradient stability is fundamentally difficult~\cite{bengio1994learning},
its effects can be mitigated through architectural variations such as LSTM~\cite{lstm}, GRU~\cite{cho2014learning}. In this works, we rely on the Batch-Normalized variant of LSTM~\cite{cooijmans2016recurrent},
that successfully applied the batch-normalization transform~\cite{batchnorm} to recurrent networks, with recurrent transition given by
\begin{eqnarray}
\left(\begin{array}{ccc}
\tilde{\vect{i}}_t \\
\tilde{\vect{f}}_t \\
\tilde{\vect{o}}_t \\
\tilde{\vect{g}}_t
\end{array}\right)
= 
 \mathrm{BN} (\mat{W}_w \vect{w}_t, \gamma_w) +
 \mathrm{BN} (\mat{W}_h \vect{h}_{t-1}, \gamma_h) +
 \vect{b},
\end{eqnarray}
where
\begin{eqnarray}
\vect{c}_t &=& \sigma(\tilde{\vect{i}}_t) \ewprod \tanh(\tilde{\vect{g}_t}) + 
               \sigma(\tilde{\vect{f}}_t) \ewprod \vect{c}_{t-1} \\
\vect{h}_t &=& \sigma(\tilde{\vect{o}}_t) \ewprod \tanh(
 \mathrm{BN} (\vect{c}_t; \gamma_c) +
 \vect{b}_c
),
\end{eqnarray}
and where  
\begin{equation}
\mathrm{BN}(\vect{x} ; \gamma) =  \gamma \ewprod
\frac{\vect{x} -   \widehat{\mathbb{E  }}[\vect{x}]}
     {       \sqrt{\widehat{\mathrm{Var}}[\vect{x}] + \epsilon}}
\end{equation}
is the  batch-normalizing transform with $\widehat{\mathbb{E}}[\vect{x}], \widehat{\mathrm{Var}}[\vect{x}]$  being the activation mean and variance estimated from the mini-batch samples. $\vect{W}_h \in \reals^{d_h \times 4 d_h}, \vect{W}_w  \in \reals^{d_w \times 4 d_h}, \vect{b} \in \reals^{4 d_h}$
and the initial states $\vect{h}_0 \in \reals^{d_h}, \vect{c}_0 \in \reals^{d_h}$
are model parameters.
$\sigma$ is the logistic sigmoid function, and the $\ewprod$ operator denotes the Hadamard product.


\subsection{Video Encoder}

We now detail the implementation of the video encoder $\Phi_v$ which extracts a fixed-length representation from the sequence of 2D frames composing a video.

Following recent work in video modelling~\cite{srivastava2015unsupervised, donahue2014long},
we leverage 2D (or 3D) convolutional neural networks that map each frame (or sequence of frames) into a
sequence vector, and then apply a recurrent neural network to extract a fixed length representations from the sequences of vectors. As for the question encoder, we rely on the Batch-Normalized LSTM~\cite{cooijmans2016recurrent} to model the sequence of vectors.

\section{Experiments and Discussion}

In this section we provide more insights about our fill-in-the-blank dataset.
We perform several experiments and explore the performance of different baseline models. We also compare our models with human performances and show that there is a significant gap to tackle.
Finally, we performs a human evaluation of our different models and shows that using the standard  metric of accuracy for comparing the different models yields results that correspond well with human assessment.

\subsection{Experimental Settings}
\label{sec:settings}


\subsubsection{Text Preprocessing}
We preprocess the questions and the answer with wordpunct tokenizer from the NLTK toolbox~\cite{nltk}.
We then lowercase all the word tokens, and end up with a vocabulary of 26,818 unique words.

\subsubsection{Video Preprocessing}
To leverage the video visual input, we investigate 2D static features and 3D moving visual features.
We rely on a GoogLeNet convolutional neural network that
has been pretrained on ImageNet~\cite{szegedy2015going}  to extract static features.
Features are extracted from the pool5/7x7 layer.
3D moving features are extracted using the C3D model~\cite{tran2015learning}, pretrained on Sport 1 million~\cite{karpathy2014large}. We apply the C3D frames on chunk of 16 consecutive frames in a video and  retrieve the activations corresponding to the ``fc7'' layer. We don't finetune the 2D and 3D CNN parameters during training on the fill-in-the-blank task.

To reduce the memory and computational requirements, we only consider a fix number
of frames/temporal segments from the different videos. If it is not specified otherwise, we consider 25 frames/temporal segments per videos. Those frames/temporal segments are sampled randomly at training while being  equally-spaced during the inference on the validation or test set.


\subsection{Language, static-visual, and moving-visual information}

\begin{table}
\caption{Fill-in-the-blank accuracy results for  single models and estimated human performance (both human experiments were conducted with a subset of 569 examples from the test set). }
  \center
  \begin{tabular}{ccc}
  \toprule
    Model & Validation & Test\\
    \midrule
    Text-only & 33.8 & 34.4\\
    GoogleNet-2D  & 34.1 & 34.9\\
    C3D       & 34.0 & 34.5\\
    GoogleNet-2D -Finetuned & 34.7 & 35.3\\
    GoogleNet-2D + C3D - Finetuned &  35.0 & 35.7\\
    \midrule
    Vocabulary* Text-only & 34.3 & 35.0\\
    Vocabulary* 2D + C3D - Finetuned & 35.4 & 36.3\\
    \midrule
    Human text-only & - & 30.2\\
    Human text+video & - & 68.7\\
    \midrule
    VGG-2D-MergingLSTMs~\cite{mazaheri2016video} & - & 34.2\\
    ResNet-2D-biLSTM-attn~\cite{challenge1} & - & 38.0\\

    \bottomrule
  \end{tabular}
\label{tab:results}
\end{table}

We test different model variations for video fill-in-the-blank based on the framework described in section~\ref{sec:framework}.
Specifically, we investigate the performance on this task of a language model only (a baseline model using only the question encoder) and  the impact of 2D and 3D features individually as well as their combination.
We train our baseline models using stochastic gradient descent along with the  Adam update rules~\cite{kingma2014adam}. Model hyperparameters can be found in the supplementary materials.

Table~\ref{tab:results} reports the valid and test accuracies for
the 5 different baseline models. 

\begin{figure}
\center
\includegraphics[width=0.45
\textwidth]{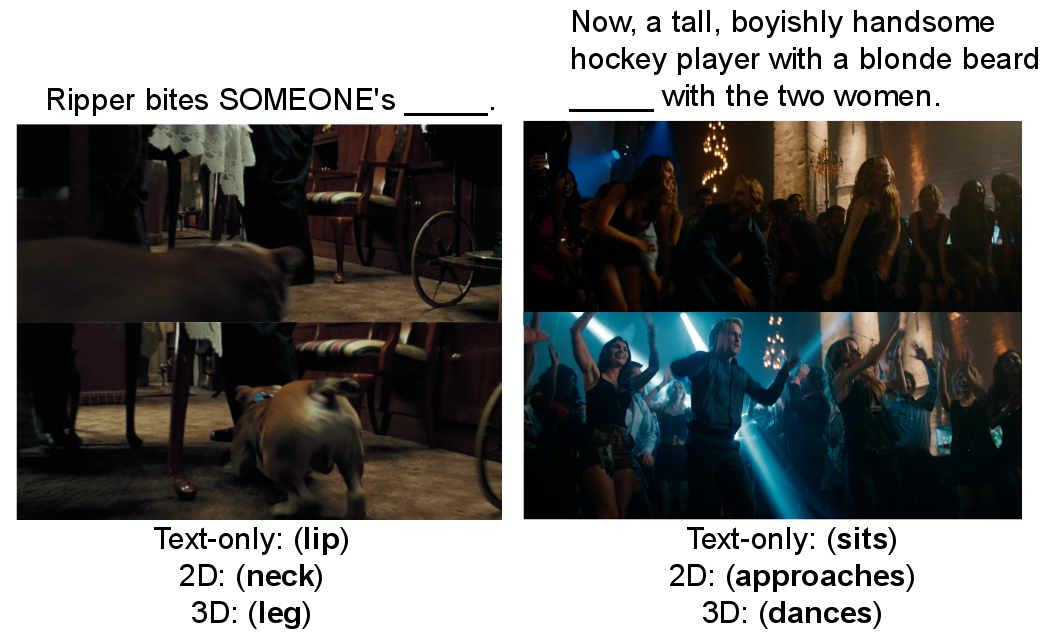}
\label{fig:qualex}
\caption{Qualitative examples for the text-only, 2D (GoogleNet-2D), and 3D (Googlenet-2D+C3D) showing the importance of visual information; in particular the importance of 3D features in recognizing actions.}
\end{figure}

While Text-only baseline obtains reasonable results
by itself, adding a visual input  to our model, through the GoogleNet-2D or C3D features,
does improve the overall accuracies. In addition; the contributions of the different
visual features seems complimentary as  they can be combined to further improve performance. To illustrate this qualitatively, in Figure \ref{fig:qualex} we show two examples which the text-only model get wrong, but which GoogleNet-2D + C3D gets right.

We also compare model with parameter initialized randomly versus model having
the text-encoder parameters initialized directly from the text-only baseline (Finetuned model in Table~\ref{tab:results}). Finetuned initialization leads to better result, we empirically observe
that it tends to reduce the model overfitting.

\begin{figure}
\center
\includegraphics[width=0.38\textwidth]{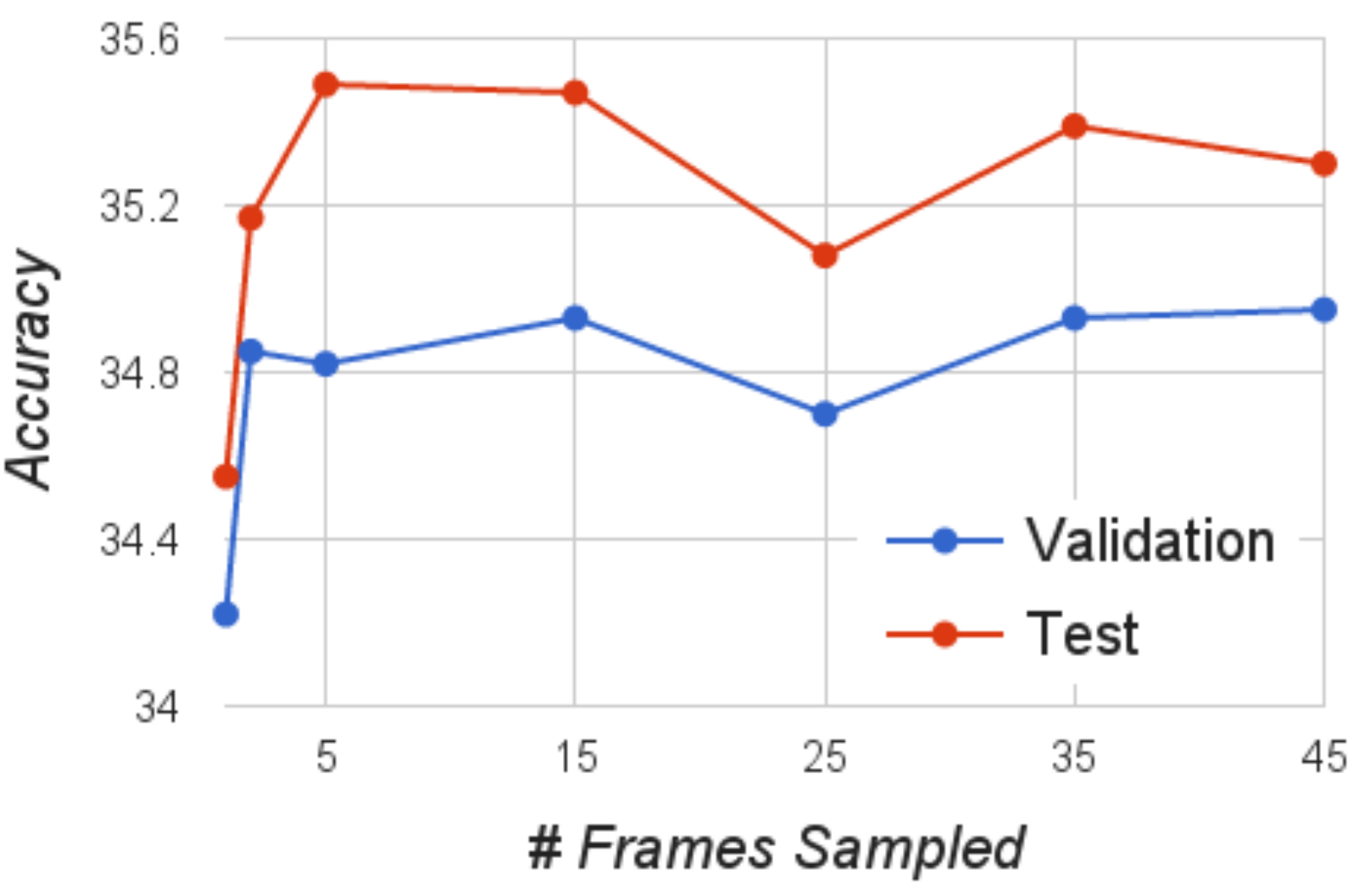}
\label{fig:frames}
\caption{Performance on the test set for GoogleNet-2D (finetuned) showing that comparable performance is achieved with just two sampled frames.}
\end{figure}

\begin{figure}
  \center
\includegraphics[width=0.48\textwidth]{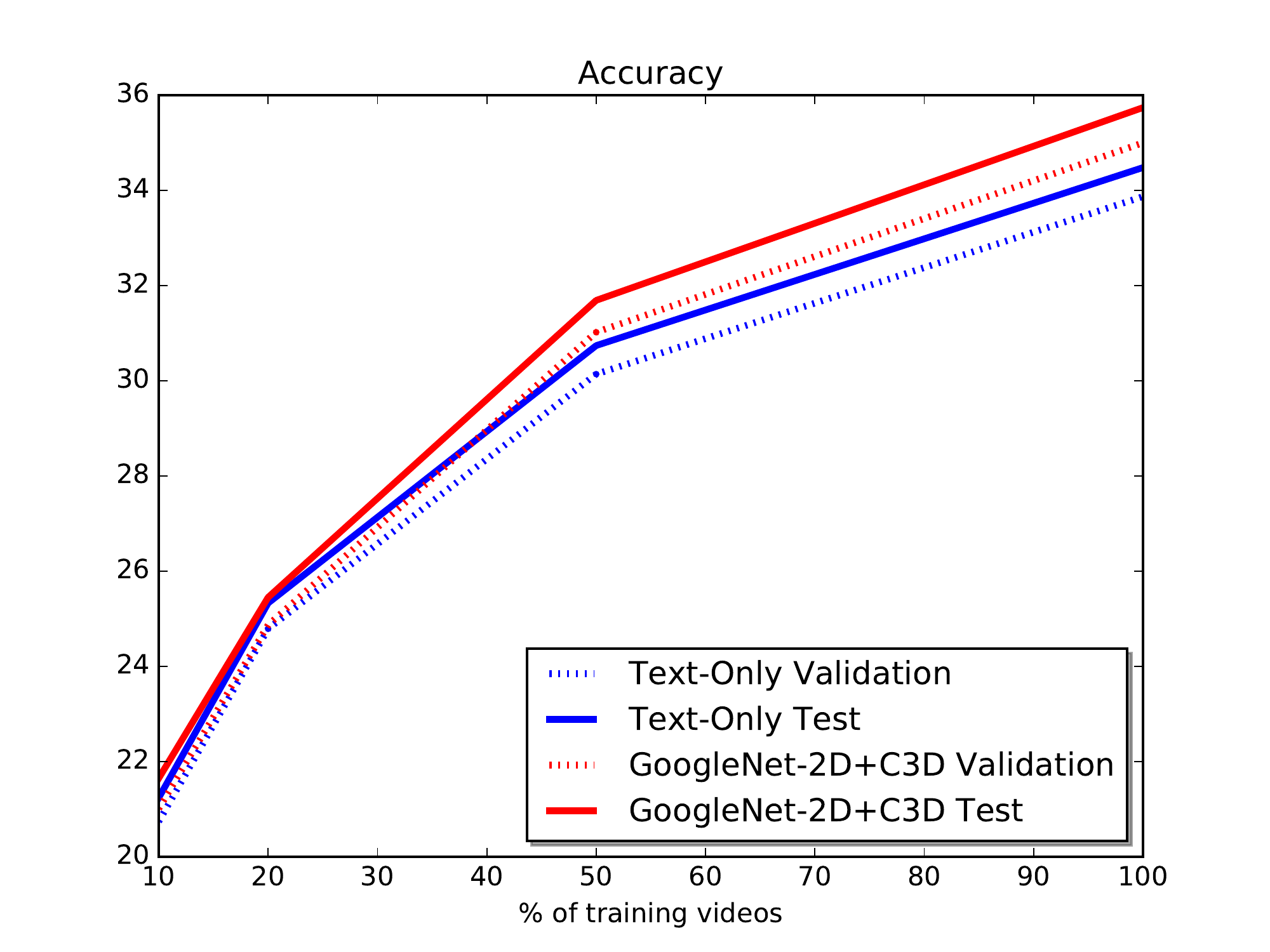}
\caption{Fill-in-the-blank accuracy results for the Text-only and GoogleNet-2D + C3D (finetuned) models on validation and test sets, trained on varying percentages (10,20,50, and 100\%) of the training data, showing a larger gain in test performance relative to validation for the video model (Note that results for models trained with 100\% of training data are the same as reported in \ref{tab:results}).}
\label{fig:data_impact}
\end{figure}

\begin{figure*}[!ht]
\center
\includegraphics[width=0.96\textwidth]{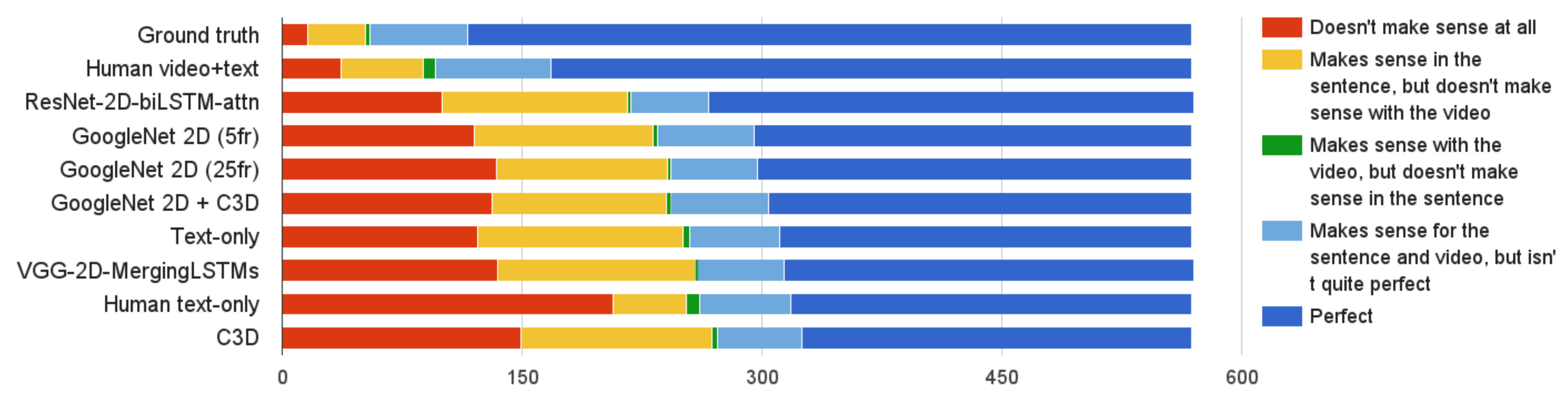}
\caption{Human evaluation of different models' answers.}
\label{fig:humaneval}
\end{figure*}

\begin{figure}
\center
\includegraphics[width=0.48\textwidth]{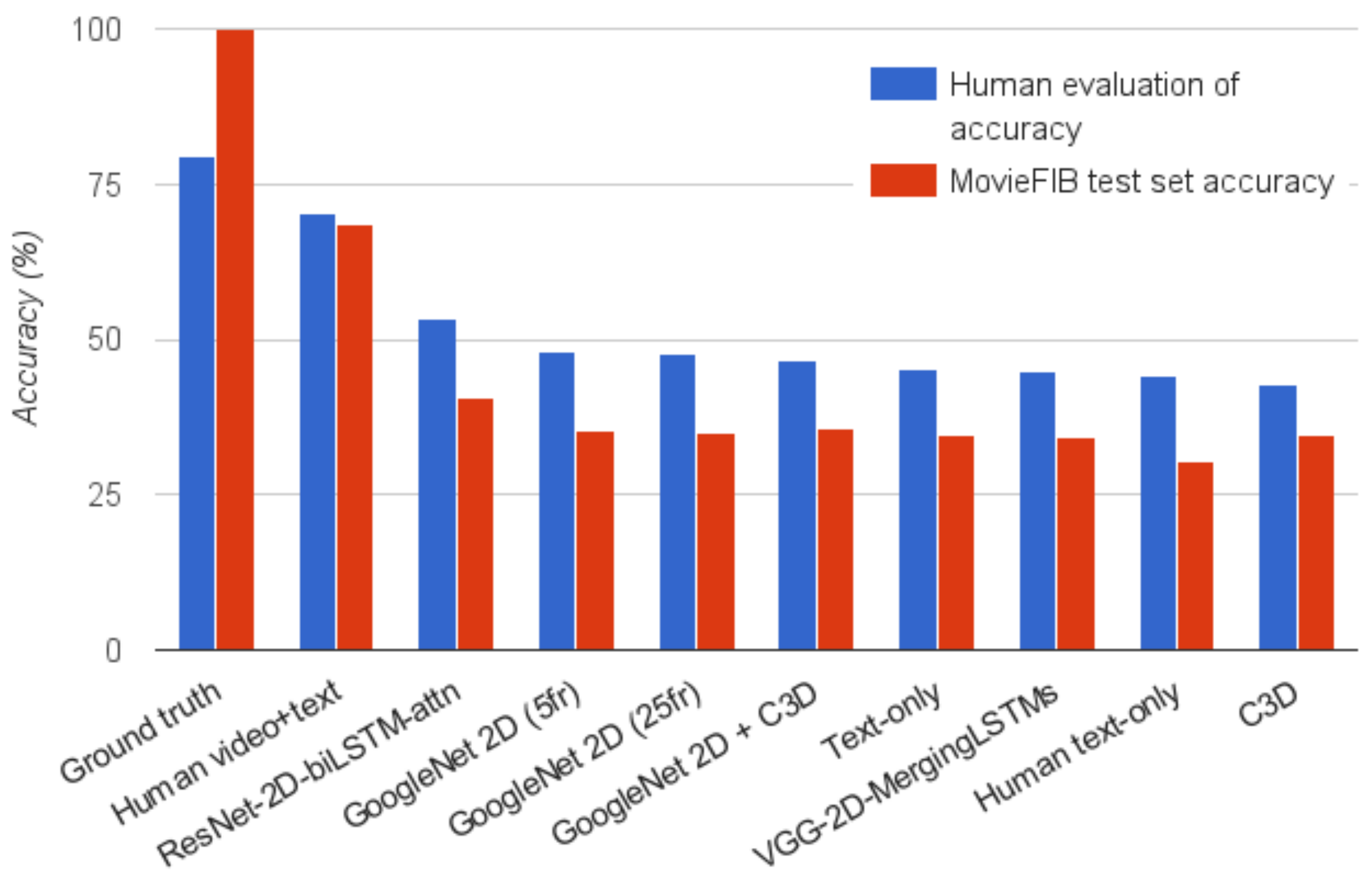}
\caption{Performance on the test set and performance according to human evaluation, demonstrating that these metrics correspond well.}
\label{fig:scatter}
\end{figure}

\subsection{Human Performances on the test set}

Table~\ref{tab:results} also reports human performance on a subset of the test set.
In order to obtain an estimate of human performance on the test set, we use Amazon Mechanical Turk to employ humans to answer a sample of 569 test examples, which is representative of the test set at a confidence of $95\%/{^+_{-}4}$. To mimic the information given to a neural network model, we require humans to fill in the blank using words from a predefined vocabulary in a searchable drop-down menu. In order to ensure quality of responses, we follow \cite{deng2009imagenet} in having 3 humans answer each question. If two or more humans answer the same for a given question, we take that as the answer; if all disagree, we randomly choose one response as the answer out of the 3 candidates. 

We perform two experiments with this setup; \textbf{human text-only} and \textbf{human text+video}. In the text-only experiment, workers are shown only the query sentence, and not the video clip, while in the text+video setting workers are given both the video clip and the query sentence.
As in the automated models, we observe that adding video input drastically improves the human performance. This confirms that visual information is of prime importance for solving this task. 

We also observe in Table~\ref{tab:results} that there is a significant gap between our
best automated model and the best human performance (on text+video), leaving some room for further improvement.
Interestingly, we notice that our text-only model outperforms the human text-only accuracy. Descriptive Video (DV) annotations are written by movie industry professionals, and have a certain linguistic style which appears to induce some statistical regularities in the text data. 
Our text-only baseline, directly trained on DV data, is able to exploit these statistical regularities, while a Mechanical Turk worker who is not familiar with the DV style of writing may miss them.

\subsection{Effect of Text and Video Preprocessing}

In this section, we report several experiments that investigate the impact
of text and video preprocessing.

We first look at the impact of the word vocabulary size. In addition to the text preprocessing
described in Section~\ref{sec:settings}, we eliminate the rare tokens from the  vocabulary applied on the input question. We only considers words that occurs more than 3 times in the training set. Rare words are replaced with an ``unknown'' token. It leads to vocabulary of size $18,663$. We also reduces the vocabulary size at the output, considering only word presents more than 50 times in the training sets,
resulting in a vocabulary of size $3,994$. This output vocabulary still includes all the
possible blanked words. We denote those variants by Vocabulary* in Table~\ref{tab:results} and observe that reducing the vocabulary size results in improved performance, highlighting the importance of the text preprocessing.


We also investigate the importance of the number of input frames for the GoogleNet-2D baseline model. Results are reported in Figure~\ref{fig:frames}. We observe that the validation performances saturates quickly, as we almost reach the best performance with only 2 sampled frames from the videos on the valid set.

\subsection{Related works using MovieFIB}
We have made the MovieFIB dataset publically available, and two recent works have made use of it. 

In \cite{challenge1}, the authors use an LSTM on pretrained ImageNet features from layer conv5b of a ResNet to encode the video, using temporal attention on frames, and a bidirectional LSTM with semantic attention for encoding the question. We refer to this model as \textbf{ResNet-2D-biLSTM-attn}, and it achieves the highest reported accuracy on our dataset so far - 38.0\% accuracy for a single model, and 40.7 for an ensemble.

In \cite{mazaheri2016video}, the authors use a similar model to our baselines, encoding video using an LSTM and pretrained VGG \cite{VGG} features, combined with the output of two LSTMs running in opposite directions on the question by an MLP. We refer to this model as \textbf{VGG-2D-MergingLSTMs}. Their method differs from ours in that they first train a Word2Vec embedding space for the questions. They find, as we do, that initializing the video encoding with the question encoding results in improved performance.

We include the results of these models in our comparisons, and report the best single-model performance of these model in \ref{tab:results}.

\begin{figure}
\center
\includegraphics[width=0.48\textwidth]{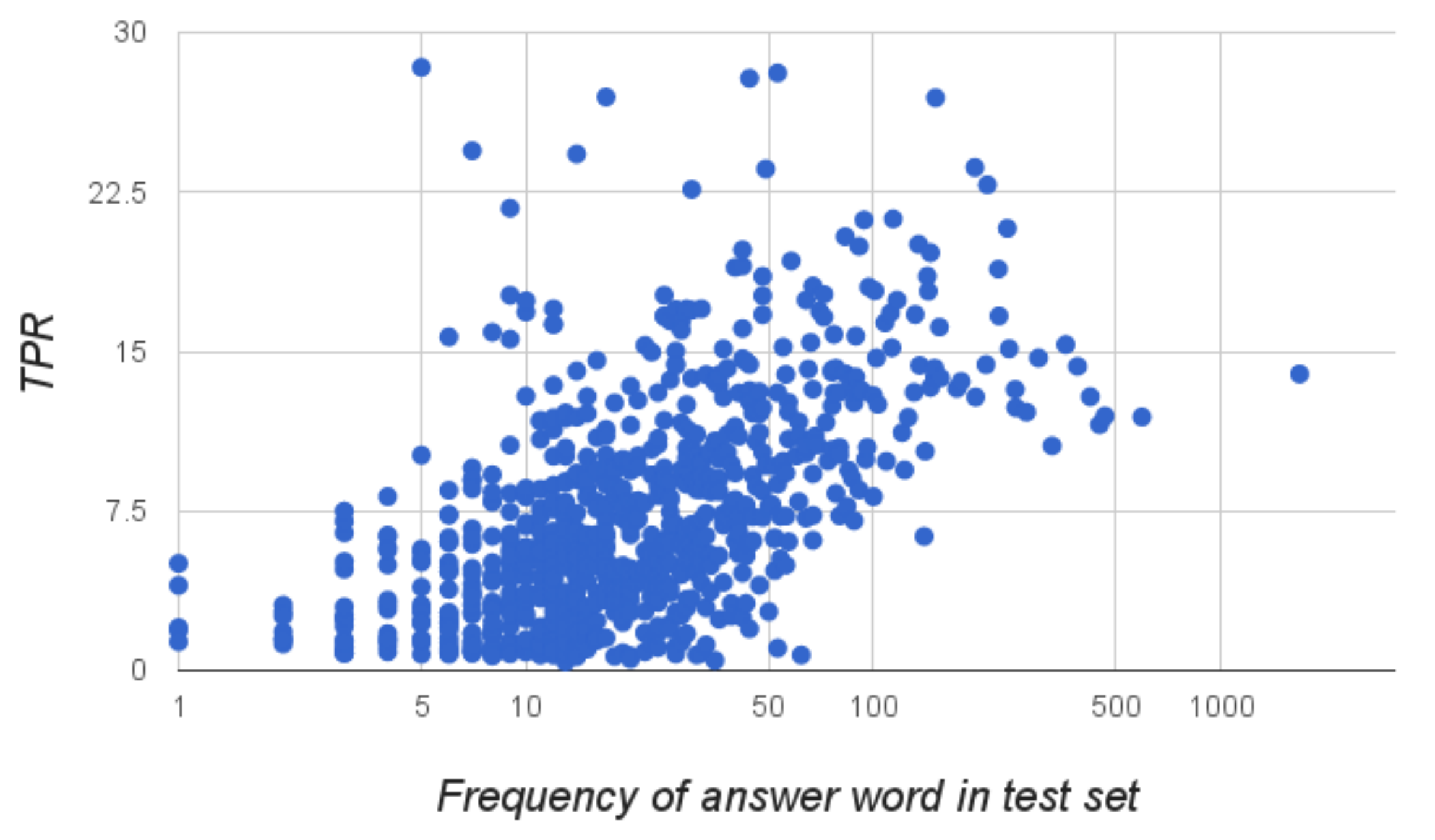}
\caption{The true positive rate (TPR) per answer word for the GoogleNet-2D+ C3D model, plotted by answer word frequency in the training set, showing that the TRP or sensitivity is highly correlated with answer word frequency.}
\label{fig:acc_by_freq}
\end{figure}

\subsection{Effects of increasing dataset size}
As evidenced by performance on large datasets like ImageNet, the amount of training data available can be a huge factor in the success of deep learning models. We are interested to know if the dataset size is an important factor in the performance of video models, and specifically, if we should expect to see an increase in the performance of existing models simply by increasing the amount of training data available.

Figure~\ref{fig:data_impact} reports the validation and test accuracies of Text-Only and GoogLeNet-2D+C3D baselines as we increase the number of training videos.
It shows that at 10\% of training data (9511 videos), text-only and video models perform very similarly ($20.7\%$ accuracy for Text-Only versus $21.0\%$ for GoogleNet-2D+C3D on the valid set).
It suggests that at 10\% of training data, there are not enough video examples for the model to leverage useful information that generalizes to unseen examples from the visual input.

However, we observe that increasing the amount of training data benefit more to the video-based model relatively to the text-only model. As data increases the performance of the video model increases more rapidly than the text-only model.
This suggests that existing video models are in fact able to gain some generalization
from the visual input given enough training examples.
Hence, Figure~\ref{fig:data_impact} highlights that further increasing the dataset size should be more beneficial for the video-based models.

Figure~\ref{fig:acc_by_freq} shows that per-word true positive rate is highly correlated with answer prevalence, indicating that increasing the number of examples for each target would likely also increase performance (we plot the results only for GoogleNet-2D + C3D for brevity, but similar correlations are seen for all models).



\subsection{Human evaluation of results}
We employ Mechanical Turk workers to rank the responses from the models described in Table \ref{tab:results}. Workers are given the clip and question, and a list of the different models' responses (including ground truth) . Figure \ref{fig:humaneval} shows how humans evaluated different models' responses. Figure \ref{fig:scatter} shows that accuracy tracks the human evaluation  well on the test set, in other words, that performance on MovieFIB is representative. Interestingly, humans evaluate that the ground truth is "Perfect" about 80\% of examples, an additional 11\% "Make sense for the sentence and video, but isn't quite perfect", and for 3\% of ground truth answers (16 examples) workers say the ground truth "Doesn't make sense at all".  We observe that for most of these examples, the issue appears to be language comprehension/style; for example "He \_\_\_\_\_ her" where the ground truth is "eyes". This may be an unfamiliar use of language for some workers. 

\section{Conclusion}

We have presented MovieFIB, a fill-in-the-blank question-answering dataset, based on descriptive video annotations for the visually impaired, with over 300,000 question-answer and video pairs.

To explore our dataset, and to better understand the capabilities of video models in general, have evaluated five different models and compared them with human performance.
In particular, we empirically observe that using visual information is of prime importance to model performance on this task, with 
our model combining 2D and  3D visual information providing the best result. However, all models still perform significantly worse than human-level, leaving room for future improvement and development of video models.

We have studied the importance of quantity of training data, showing that models leveraging visual input benefit more than text-only models from an increase of the training samples. This suggests that
performance could be further improved just by increasing the amount of training data.

Finally we have performed a human evaluation of our different models, showing that accuracy is a robust metric, corresponding well with human assessment for the fill-in-the-blank task.

We hope that the MovieFIB dataset we have introduced here will be useful to develop and evaluate models which better understand the visual content of videos, and that it will encourage further research and progress in this field.

{\small
\bibliographystyle{ieee}
\bibliography{egbib}

\begin{thebibliography}{10}\itemsep=-1pt

\bibitem{TRECVIDMED14}
Trecvid med 14.
\newblock \url{http://nist.gov/itl/iad/mig/med14.cfm}.
\newblock Accessed: 2016-11-13.

\bibitem{nltk}
S.~B. aand Edward~Loper and E.~Klein.
\newblock {\em Natural Language Processing with Python}.
\newblock O'Reilly Media Inc., 2009.

\bibitem{VQA}
S.~Antol, A.~Agrawal, J.~Lu, M.~Mitchell, D.~Batra, C.~L. Zitnick, and
  D.~Parikh.
\newblock Vqa: Visual question answering.
\newblock In {\em ICCV}, 2015.

\bibitem{Nicolas15}
N.~Ballas, L.~Yao, C.~Pal, and A.~Courville.
\newblock Delving deeper into convolutional networks for learning video
  representations.
\newblock {\em ICLR}, 2016.

\bibitem{bengio1994learning}
Y.~Bengio, P.~Simard, and P.~Frasconi.
\newblock Learning long-term dependencies with gradient descent is difficult.
\newblock {\em Neural Networks, IEEE Transactions on}, 1994.

\bibitem{cho2014learning}
K.~Cho, B.~Van~Merri{\"e}nboer, C.~Gulcehre, D.~Bahdanau, F.~Bougares,
  H.~Schwenk, and Y.~Bengio.
\newblock Learning phrase representations using rnn encoder-decoder for
  statistical machine translation.
\newblock {\em arXiv preprint arXiv:1406.1078}, 2014.

\bibitem{cooijmans2016recurrent}
T.~Cooijmans, N.~Ballas, C.~Laurent, {\c{C}}.~G{\"u}l{\c{c}}ehre, and
  A.~Courville.
\newblock Recurrent batch normalization.
\newblock {\em arXiv preprint arXiv:1603.09025}, 2016.

\bibitem{deng2009imagenet}
J.~Deng, W.~Dong, R.~Socher, L.-J. Li, K.~Li, and L.~Fei-Fei.
\newblock Imagenet: A large-scale hierarchical image database.
\newblock In {\em CVPR}, 2009.

\bibitem{denkowski2014meteor}
M.~Denkowski and A.~Lavie.
\newblock Meteor universal: Language specific translation evaluation for any
  target language.
\newblock In {\em Ninth Workshop on Statistical Machine Translation}, 2014.

\bibitem{donahue2014long}
J.~Donahue, L.~Hendricks, S.~Guadarrama, M.~Rohrbach, S.~Venugopalan,
  K.~Saenko, and T.~Darrell.
\newblock Long-term recurrent convolutional networks for visual recognition and
  description.
\newblock {\em arXiv preprint arXiv:1411.4389}, 2014.

\bibitem{Guadarrama_youtube2text}
S.~Guadarrama, N.~Krishnamoorthy, G.~Malkarnenkar, S.~Venugopalan, R.~Mooney,
  T.~Darrell, and K.~Saenko.
\newblock Youtube2text: Recognizing and describing arbitrary activities using
  semantic hierarchies and zero-shot recognition.
\newblock In {\em ICCV}, 2013.

\bibitem{hochreiter1991untersuchungen}
S.~Hochreiter.
\newblock Untersuchungen zu dynamischen neuronalen netzen.
\newblock {\em Master's thesis}, 1991.

\bibitem{lstm}
S.~Hochreiter and J.~Schmidhuber.
\newblock Long short-term memory.
\newblock {\em Neural computation}, 1997.

\bibitem{batchnorm}
S.~Ioffe and C.~Szegedy.
\newblock Batch normalization: Accelerating deep network training by reducing
  internal covariate shift.
\newblock {\em ICML}, 2015.

\bibitem{karpathy2014large}
A.~Karpathy, G.~Toderici, S.~Shetty, T.~Leung, R.~Sukthankar, and L.~Fei-Fei.
\newblock Large-scale video classification with convolutional neural networks.
\newblock In {\em CVPR}, 2014.

\bibitem{kingma2014adam}
D.~Kingma and J.~Ba.
\newblock Adam: A method for stochastic optimization.
\newblock {\em arXiv preprint arXiv:1412.6980}, 2014.

\bibitem{kojima2002}
A.~Kojima, T.~Tamura, and K.~Fukunaga.
\newblock Natural language description of human activities from video images
  based on concept hierarchy of actions.
\newblock {\em IJCV}, 2002.

\bibitem{krizhevsky2012imagenet}
A.~Krizhevsky, I.~Sutskever, and G.~E. Hinton.
\newblock Imagenet classification with deep convolutional neural networks.
\newblock In {\em NIPS}, 2012.

\bibitem{lin2004rouge}
C.-Y. Lin.
\newblock Rouge: A package for automatic evaluation of summaries.
\newblock In {\em Text summarization branches out: Proceedings of the ACL-04
  workshop}. Barcelona, Spain, 2004.

\bibitem{lin2014microsoft}
T.-Y. Lin, M.~Maire, S.~Belongie, J.~Hays, P.~Perona, D.~Ramanan,
  P.~Doll{\'a}r, and C.~L. Zitnick.
\newblock Microsoft coco: Common objects in context.
\newblock In {\em ECCV}, 2014.

\bibitem{maxentNLTK}
G.~Malecha and I.~Smith.
\newblock Large scale movie description and understanding challenge, 2010.

\bibitem{malinowski2014multi}
M.~Malinowski and M.~Fritz.
\newblock A multi-world approach to question answering about real-world scenes
  based on uncertain input.
\newblock In {\em NIPS}, 2014.

\bibitem{mazaheri2016video}
A.~Mazaheri, D.~Zhang, and M.~Shah.
\newblock Video fill in the blank with merging lstms.
\newblock {\em arXiv preprint arXiv:1610.04062}, 2016.

\bibitem{papineni2002bleu}
K.~Papineni, S.~Roukos, T.~Ward, and W.-J. Zhu.
\newblock Bleu: a method for automatic evaluation of machine translation.
\newblock In {\em ACL}, 2002.

\bibitem{maxentRatnaparkhi1996}
A.~Ratnaparkhi.
\newblock A maximum entropy model for part-of-speech tagging.
\newblock In {\em {EMNLP}}. 1996.

\bibitem{regneri2013grounding}
M.~Regneri, M.~Rohrbach, D.~Wetzel, S.~Thater, B.~Schiele, and M.~Pinkal.
\newblock Grounding action descriptions in videos.
\newblock {\em ACL}, 2013.

\bibitem{ren2015exploring}
M.~Ren, R.~Kiros, and R.~Zemel.
\newblock Exploring models and data for image question answering.
\newblock In {\em NIPS}, 2015.

\bibitem{rohrbach15cvpr}
A.~Rohrbach, M.~Rohrbach, N.~Tandon, and B.~Schiele.
\newblock A dataset for movie description.
\newblock In {\em CVPR}, 2015.

\bibitem{rohrbach2015dataset}
A.~Rohrbach, M.~Rohrbach, N.~Tandon, and B.~Schiele.
\newblock A dataset for movie description.
\newblock In {\em CVPR}, pages 3202--3212, 2015.

\bibitem{lsmdc-challenge}
A.~Rohrbach, A.~Torabi, T.~Maharaj, M.~Rohrbach, C.~Pal, A.~Courville, and
  B.~Schiele.
\newblock Large scale movie description and understanding challenge, 2016.

\bibitem{lsmdc2015}
A.~Rohrbach, A.~Torabi, M.~Rohrbach, N.~Tandon, P.~Chris, L.~Hugo, C.~Aaron,
  and B.~Schiele.
\newblock Movie description.
\newblock {\em arXiv preprint}, 2016.

\bibitem{rohrbach2013}
M.~Rohrbach, W.~Qiu, I.~Titov, S.~Thater, M.~Pinkal, and B.~Schiele.
\newblock Translating video content to natural language descriptions.
\newblock In {\em ICCV}, 2013.

\bibitem{silberman2012indoor}
N.~Silberman, D.~Hoiem, P.~Kohli, and R.~Fergus.
\newblock Indoor segmentation and support inference from rgbd images.
\newblock In {\em ECCV}, 2012.

\bibitem{VGG}
K.~Simonyan and A.~Zisserman.
\newblock Very deep convolutional networks for large-scale image recognition.
\newblock {\em arXiv preprint arXiv:1409.1556}, 2014.

\bibitem{srivastava2015unsupervised}
N.~Srivastava, E.~Mansimov, and R.~Salakhutdinov.
\newblock Unsupervised learning of video representations using lstms.
\newblock In {\em ICML}, 2015.

\bibitem{sutskever2014sequence}
I.~Sutskever, O.~Vinyals, and Q.~V. Le.
\newblock Sequence to sequence learning with neural networks.
\newblock In {\em NIPS}, 2014.

\bibitem{szegedy2015going}
C.~Szegedy, W.~Liu, Y.~Jia, P.~Sermanet, S.~Reed, D.~Anguelov, D.~Erhan,
  V.~Vanhoucke, and A.~Rabinovich.
\newblock Going deeper with convolutions.
\newblock In {\em CVPR}, 2015.

\bibitem{MovieQA}
M.~Tapaswi, Y.~Zhu, R.~Stiefelhagen, A.~Torralba, R.~Urtasun, and S.~Fidler.
\newblock Movieqa: Understanding stories in movies through question-answering.
\newblock In {\em CVPR}, 2016.

\bibitem{AtorabiM-VAD2015}
A.~Torabi, C.~Pal, H.~Larochelle, and A.~Courville.
\newblock Using descriptive video services to create a large data source for
  video annotation research.
\newblock {\em arXiv preprint}, 2015.

\bibitem{tran2015learning}
D.~Tran, L.~Bourdev, R.~Fergus, L.~Torresani, and M.~Paluri.
\newblock Learning spatiotemporal features with 3d convolutional networks.
\newblock In {\em ICCV}, 2015.

\bibitem{vedantam2015cider}
R.~Vedantam, C.~Lawrence~Zitnick, and D.~Parikh.
\newblock Cider: Consensus-based image description evaluation.
\newblock In {\em CVPR}, 2015.

\bibitem{venugopalan2014translating}
S.~Venugopalan, H.~Xu, J.~Donahue, M.~Rohrbach, R.~Mooney, and K.~Saenko.
\newblock Translating videos to natural language using deep recurrent neural
  networks.
\newblock {\em NAACL}, 2015.

\bibitem{yao2015describing}
L.~Yao, A.~Torabi, K.~Cho, N.~Ballas, C.~Pal, H.~Larochelle, and A.~Courville.
\newblock Describing videos by exploiting temporal structure.
\newblock In {\em ICCV}, 2015.

\bibitem{challenge1}
Y.~Yu, H.~Ko, J.~Choi, and G.~Kim.
\newblock Video captioning and retrieval models with semantic attention.
\newblock {\em arXiv preprint arXiv:1610.02947}, 2016.

\bibitem{zhu2015uncovering}
L.~Zhu, Z.~Xu, Y.~Yang, and A.~G. Hauptmann.
\newblock Uncovering temporal context for video question and answering.
\newblock {\em arXiv preprint arXiv:1511.04670}, 2015.

\bibitem{zitnick2016adopting}
C.~L. Zitnick, R.~Vedantam, and D.~Parikh.
\newblock Adopting abstract images for semantic scene understanding.
\newblock {\em PAMI}, 2016.

\end{thebibliography}
}

\end{document}